\documentclass[11pt,a4paper]{article}
\usepackage[hyperref]{acl2020}
\usepackage{times}
\usepackage{latexsym}

\usepackage[utf8]{inputenc}
\usepackage[english]{babel}

\usepackage{array, makecell}
\usepackage{amsmath,amsthm}
\usepackage{multirow}
\usepackage{microtype}
\usepackage{booktabs}

\newcommand\blfootnote[1]{%
  \begingroup
  \renewcommand\thefootnote{}\footnote{#1}%
  \addtocounter{footnote}{-1}%
  \endgroup
}

\usepackage{tikz}

\usepackage[noend]{algorithmic}
\usepackage{algorithm}

\usepackage{microtype}

\aclfinalcopy 

\title{Studying Strategically: Learning to Mask for Closed-book QA}

\author{Qinyuan Ye$^{1\dagger}$ \quad Belinda Z. Li$^{2\ddagger}$ \quad Sinong Wang$^{3}$ \quad Benjamin Bolte$^{3}$\\\textbf{Hao Ma$^{3}$  \quad Wen-tau Yih$^{3}$ \quad Xiang Ren$^{1}$ \quad Madian Khabsa$^{3}$}\\
$^{1}$University of Southern California \quad $^{2}$MIT CSAIL \quad $^{3}$Facebook AI\\
\texttt{\{qinyuany,xiangren\}@usc.edu} \quad \texttt{bzl@mit.edu}\\
\texttt{\{sinongwang,bbolte,scottyih,haom,mkhabsa\}@facebook.com}
}

\date{}

\begin{document}
\maketitle
\begin{abstract}
Closed-book question-answering (QA) is a challenging task that requires a model to directly answer questions without access to external knowledge.
It has been shown that directly fine-tuning pre-trained language models with (question, answer) examples yields surprisingly competitive performance,
which is further improved upon through adding an
intermediate
pre-training stage between general pre-training and fine-tuning.
Prior work used a heuristic during this intermediate stage, whereby named entities and dates are masked, and the model is trained to recover these tokens. 
In this paper, we aim to \textit{learn} the optimal masking strategy for the intermediate pre-training stage.
We first train our masking policy to extract spans that are likely to be tested, using supervision from the downstream task itself, then deploy the learned policy during intermediate pre-training. 
Thus, our policy packs \textit{task-relevant} knowledge into the parameters of a language model. 
Our approach is particularly effective on TriviaQA, outperforming strong heuristics 
when used to pre-train BART.




\end{abstract}

\section{Introduction}


Prior work has shown that large, neural language models (LMs) have a remarkable capability to learn knowledge from pre-training, which is stored implicitly within their parameters \cite{petroni-etal-2019-language}.\blfootnote{$^\dagger$Work partially done while interning at Facebook AI.}
This encoded knowledge allows LMs to perform a wide variety
of knowledge-intensive tasks \textit{without} an external knowledge base, including 
knowledge base relation inference and fact checking~\cite{petroni-etal-2019-language, lee-etal-2020-language}.
The implicit knowledge capacity can also be tested in the form of question answering, in the task termed as ``Closed-book QA" \cite{roberts2020much}.\blfootnote{$^\ddagger$Work partially done while working at Facebook AI.} By directly fine-tuning the LM with (question, answer) pairs, closed-book methods perform competitively with retrieval-based methods on open-domain QA benchmarks.\footnote{Extremely large language models such as GPT-3~\cite{brown2020language} performed competitively on closed-book QA benchmarks without any fine-tuning (zero-shot).} 

\begin{figure}
    \centering
    \includegraphics[trim=0.3cm 0.3cm 0.2cm 0,clip,width=0.5\textwidth]{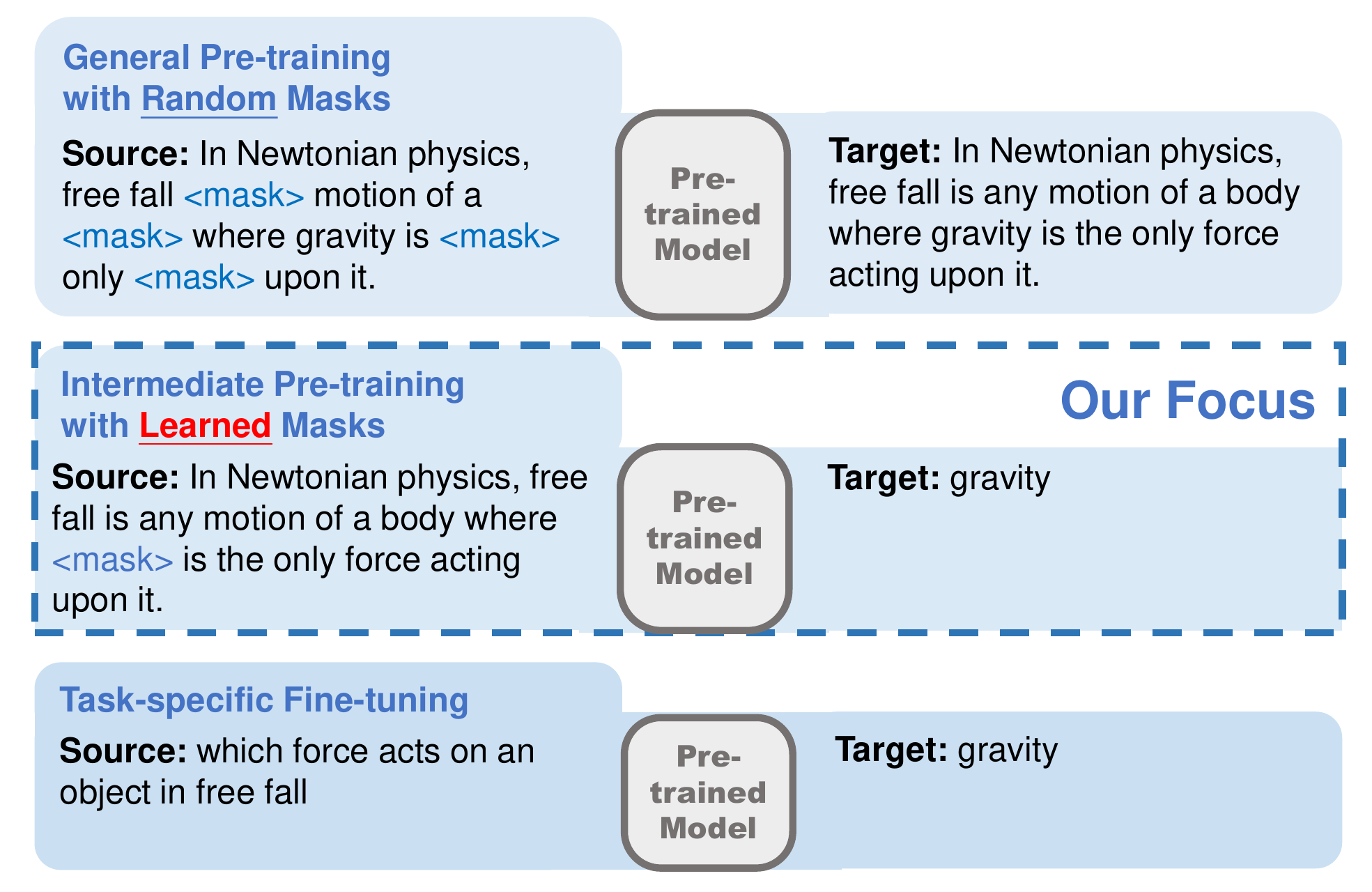}
    \caption{\textbf{Motivating Example.} Intermediate pre-training using salient span masking is helpful in improving closed-book QA performance \cite{roberts2020much}. We propose to use a \textit{learned} masking policy during intermediate pre-training to pack more task-relevant knowledge into the parameters of LM.}
    \label{fig:intro}
    \vspace{-0.5cm}
\end{figure}

One integral component of the success of closed-book QA is \textit{salient span masking} (SSM, \citealt{guu2020realm}), an intermediate pre-training task that bridges general pre-training and task-specific fine-tuning by training the model to recover masked named entities or dates in the pre-training corpus (See Fig.~\ref{fig:intro} for illustration). 
Notably, SSM brings 9\%+ EM improvement on TriviaQA \cite{joshi-etal-2017-triviaqa} when T5-11B is used. For comparison, scaling T5 model from 3B parameters to 11B only yielded 7\% improvements~---~indicating that a good choice of masking strategy could be even more influential than scaling the model size.

This observation leads to our research question~---~How can we find a policy that generates masks more strategically? Such masking policy will pack more \textit{task-relevant} knowledge into the LM, and subsequently provide a better initialization for fine-tuning on closed-book QA tasks. In short, building upon 
``\textit{How much} knowledge can you pack into the parameters of a language model?'' 
\cite{roberts2020much}, we seek to answer ``\textit{What} knowledge do you want to pack into the parameters of a language model?''

To this end, we propose to learn a masking policy that models which spans in the pre-training corpus are likely to be queried, and thus, should be masked. Given a question $q$, an answer span~$a$, and the context passage $c$ that the answer is extracted from (usually given in the dataset), 
we learn to extract answer span $a$ from context paragraph $c$, without prior knowledge of what the question $q$ is. 
This masking policy is analogous to the ``gap selection'' model in question generation tasks \cite{becker-etal-2012-mind}. It is expected to capture a student's meta-knowledge when preparing for a closed-book exam: given a textbook, the student knows what contents are likely to be tested (presumably from previous experience of taking exams), and thus the student will try to focus on these contents.

We use BART-base \cite{lewis-etal-2020-bart} as our backbone and extensively experiment with several masking policies learned from different sources of supervision. With a masking policy learned on TriviaQA, we achieve $24.71\%_{(\pm .21)}$ exact match on TriviaQA, surpassing $23.62\%_{(\pm .29)}$ using SSM and $22.93\%_{(\pm.14)}$ using random masking. The same trend holds when we deploy our best-performing policy to BART-large. We also observe that learned masking policies can positively transfer in some (but not all) cases: in these cases, a policy learned from one QA dataset can benefit other QA datasets.


\section{Preliminaries}
Masked language modeling (MLM) is a widely-used self-supervised pre-training objective \cite{devlin-etal-2019-bert}. MLM and its variants can be characterized with two key components: a masking policy $g(.;\phi)$, parameterized by $\phi$, which decides the collection of tokens to be masked, and a language model $f(.;\theta)$, parameterized by $\theta$.
Formally, given $\mathbf{x}=[x_1, x_2, ..., x_m]$, a sequence of tokens in the pre-training corpus, $g(\mathbf{x};\phi)$ generates a sequence of binary decisions $\mathbf{d}=[d_1, d_2, ..., d_m]$, where $d_i=1$ indicates the token $x_i$ will be masked. In one pre-training update, the masking policy first computes a corrupted input $\mathbf{x}'$ , where $x'_i=x_i$ if $d_i=0$ and $x'_i=\texttt{<mask>}$ if $d_i=1$. The language model $f(.;\theta)$ then learns to generate the masked tokens from this corrupted input. Following SSM, we opt to mask and reconstruct exactly one span, so $g(\mathbf{x};\phi)$ can take an equivalent form of predicting the start and end positions of a span. 

\begin{figure}
    \centering
    \includegraphics[trim=0.7cm 0.4cm 0.3cm 0.2cm,clip,width=0.48\textwidth]{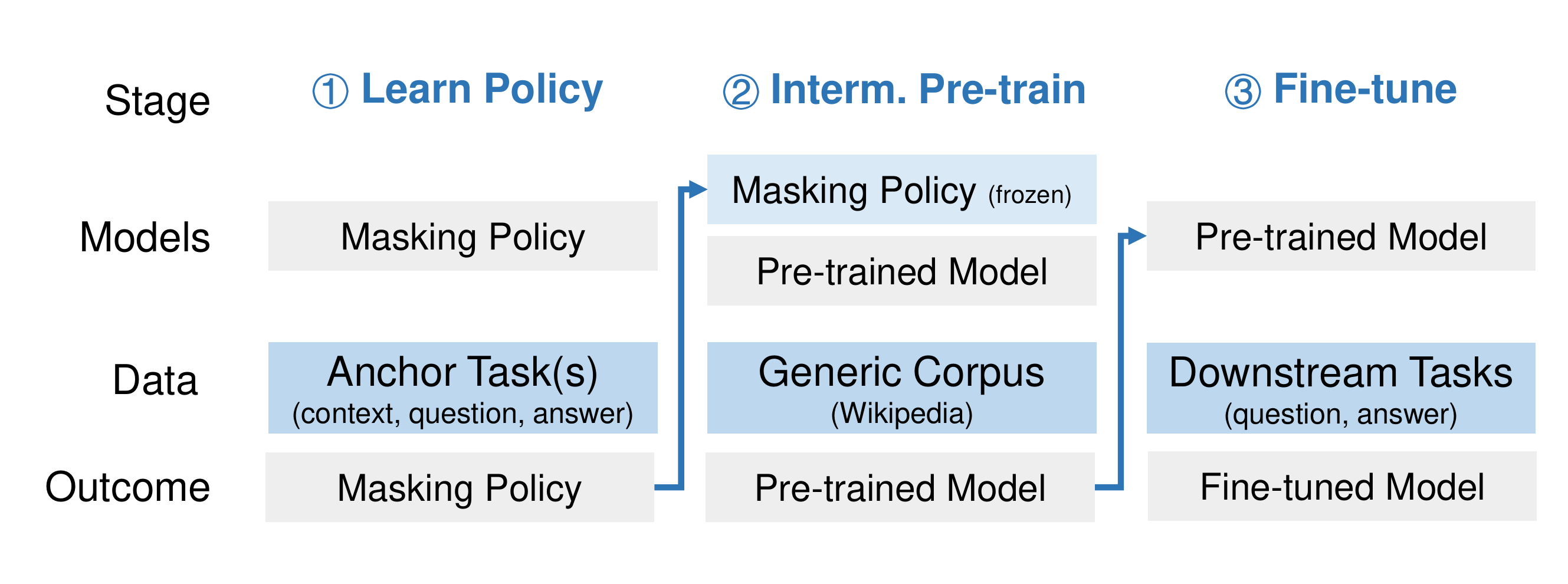}
    \vspace{-0.6cm}
    \caption{\textbf{Stages in the proposed work.} \textbf{(1) Learn a Masking Policy:} Learn a masking policy on one or a few anchor tasks; the policy learns to mask important information in the context. \textbf{(2) Intermediate Pre-train:} Pre-train the language model by using the learned policy to mask generic corpus, thus knowledge of interest is implicitly encoded into the LM. \textbf{(3) Fine-tune:} Fine-tune a pre-trained checkpoint on downstream closed-book QA tasks.}
    \vspace{-0.4cm}
    \label{fig:concept}
\end{figure}

Our goal is to learn a masking policy $g(.;\phi)$ that helps ``pack" task-relevant knowledge into LM parameters (Stage~1 in Fig.~\ref{fig:concept}). To effectively learn the masking policy, we assume access to (context, question, answer) examples for at least one QA dataset\footnote{The context can be annotated by humans (Natural Questions) or obtained from distant supervision (TriviaQA). The latter requires simple retrieval methods such as BM25.}, which we refer to as \textit{anchor task(s)}. 
To evaluate a learned policy, we deploy it in intermediate pre-training (Stage 2), on a preset corpus (\textit{i.e.}, Wikipedia), then fine-tune on downstream closed-book QA tasks (Stage 3). 

\section{Approach}
\paragraph{Overview.} Given the triplet (context, question, answer), we learn a masking policy from (context, answer) pairs only, training our policy to extract the \textit{answer} within the \textit{context}. At a high level, we want to learn to mask \textit{likely answers} from an evidence document, such that during the pre-training phase, the language model will focus on ``memorizing", or learning to unmask, spans that resemble answers.

Since the policy will be deployed to make decisions on a large-scale corpus, we opt to use light-weight architectures instead of larger pre-trained models. We encode the context sequence with a 2-layer Bi-LSTM model, and then use a linear layer to predict the start and end position of a potential answer span.

\paragraph{Model.} More specifically, given context paragraph tokens $\mathbf{x}=[x_1, x_2, ..., x_m]$, we first use an embedding matrix $\mathbf{E}$ to embed each token: $[\mathbf{e}_1, \mathbf{e}_2, ..., \mathbf{e}_n]$. Then, we use a 2-layer bidirectional LSTM model to compute the hidden representation at each position. 
\vspace{-0.1cm}
\begin{equation}
    [\mathbf{h}_1, \mathbf{h}_2, ..., \mathbf{h}_n] = \textrm{Bi-LSTM}([\mathbf{e}_1, \mathbf{e}_2, ..., \mathbf{e}_n])
\end{equation}

Finally, we use two learned vectors $(\mathbf{w}_{st}, b_{st})$ and $(\mathbf{w}_{ed}, b_{ed})$ to compute the logits for each position being the start or end position of the potential answer span. For example, the logit of position $j$ being a start/end position is computed as follows.
\vspace{-0.1cm}
\begin{equation}
\begin{aligned}
    y_{j, st} = \mathbf{w}_{st} \mathbf{h}_j + b_{st} \\
    y_{j, ed} = \mathbf{w}_{ed} \mathbf{h}_j + b_{ed}
\end{aligned}
\end{equation}

\paragraph{Policy Inference.} When deploying the policy to intermediate pre-training, we select the potential answer spans by ranking the sum of start and end logits of each potential spans, in accordance to the inference step in machine reading comprehension models. That is, we rank the spans $(i,j)$ according to $y_{i, st}+y_{j, ed}$. We consider two variants when deploying the policy: (a) masking the top 1 span or (2) sampling 1 span from the top 5 spans.

\paragraph{Anchor Tasks.} We consider three sources of supervision for our policy. (1) Natual Questions (NQ, \citealt{kwiatkowski-etal-2019-natural}); (2) TriviaQA (TQA, \citealt{joshi-etal-2017-triviaqa}); (3) a mixture of NQ, TQA, SQuAD \cite{rajpurkar-etal-2016-squad} and ReCoRD \cite{Zhang2018ReCoRDBT} datasets. For NQ and TQA, we directly use the (context, question, answer) training data released by DPR~\cite{karpukhin-etal-2020-dense}. SQuAD and ReCoRD are extractive reading comprehension datasets, so that (context, question, answer) examples can be directly obtained. 

\paragraph{Training Details.} The embedding matrix $\mathbf{E}$ is initialized with the weights in BART-base model. We optimize cross entropy loss between the logits outputted by the model and the gold annotations. For each source of supervision stated above, we train the policy for 30 epochs with learning rate of 1e-5 and batch size of 512, and select the best checkpoint according to validation loss.

\section{Experiments}

\subsection{Experiment Setup}
\paragraph{Datasets.}
Following \citet{roberts2020much}, we experiment with three open-domain QA benchmarks: Natural Questions (NQ), WebQuestions (WQ, \citealt{berant-etal-2013-semantic}) and \mbox{TriviaQA} (TQA) in the closed-book setting. We use the train/dev/test splits that are consistent with \citet{lee-etal-2019-latent} and \citet{karpukhin-etal-2020-dense}. 

\paragraph{Baselines.}
We report closed-book QA performance when fine-tuned on the following different checkpoints: (1) Publicly released BART-base model (BART); (2) Intermediate pre-training by masking $15\%$ randomly selected tokens (+Random); (3) Intermediate pre-training with salient span masking\footnote{The named entity tags are obtained with \href{https://spacy.io/}{spaCy}.} \cite{roberts2020much, guu2020realm} (+SSM). Alongside, we report performance when applying our own learned policy to intermediate pre-training (+Learned). To ensure fair comparison, all intermediate pre-training is done with input sequence length of 128, batch size of 2048, learning rate of $0.0001$, and a total number of $100,000$ updates, using Wikipedia snapshot from December 20, 2018\footnote{The snapshot is available at \url{https://archive.org/download/enwiki-20181220}. Wikipedia is licensed under \href{https://creativecommons.org/licenses/by-sa/3.0/}{CC BY-SA 3.0}.}. We use the linear learning rate warmup schedule for the first $6\%$ of steps, and linear decay for the remaining steps.

\paragraph{Fine-tuning.}
We take each checkpoint from the baselines, along with the checkpoint using our own learned policy, and fine-tune it on the three closed-book QA datasets separately. We use Adam \cite{Kingma2015AdamAM} optimizer for BART-base experiments, and Adafactor~\cite{pmlr-v80-shazeer18a} for BART-large. For hyperparameter settings, please refer to Appendix \ref{app:hps}. We report the average and standard deviation of performance using three random seeds. We report exact match after standard normalization \cite{rajpurkar-etal-2016-squad}.

\subsection{Results and Discussion} 

\paragraph{TriviaQA Results.} We first experiment with TriviaQA since TriviaQA benefits most from SSM in \cite{roberts2020much}. That is, we learn a masking policy from TriviaQA, plug the policy to intermediate pre-training, and then fine-tune on TriviaQA. We list the performance in Table \ref{tab:closed-book-tqa}. 
From the table, we make multiple key observations.

First of all, we observe performance gain with further pre-training with random masks on BART-base. This may be due to the common observation that language models tend to improve with further pretraining even after validation perplexity have plateaued \cite{conneau2019unsupervised}, or that Wikipedia corpus is more closely related to the three closed-book QA tasks\footnote{BART was originally pre-trained on a combination of news, books, stories, and web text, the same as in RoBERTa \cite{liu2019roberta}.}.

In addition, SSM does indeed serve as a strong heuristic objective as described in \citet{roberts2020much}. Our absolute performance gain is less significant, with $0.8\%$ improvement on BART-base and $2.0\%$ improvement on BART-large. We hypothesize this is due to the smaller size of BART models (140M/406M parameters for BART-base/large, compared to 11B in T5-11B).

As for our proposed approach, the masking policy learned on TriviaQA achieves the strongest performance, surpassing SSM in both BART-base and BART-large experiments. This demonstrates that the our method is effective for TriviaQA. Notably, pre-training BART-base with the learned policy outperforms original BART-large model (EM=$24.71\%$ vs EM=$24.28\%$, respectively), effectively saving 260M parameters in the LM. 
\begin{table}[t]
\centering
\scalebox{0.75}{
\begin{tabular}{lcccccc}
\toprule
Model (\#param)     &  Dev & Test \\
\midrule
BART-base (140M)& $22.34_{\pm.31}$ & $21.82_{\pm.15}$   \\
\quad +Random &  $23.02_{\pm.24}$ & $22.93_{\pm.14}$   \\
\quad +SSM &  $24.53_{\pm.14}$ & $23.62_{\pm.29}$    \\
\quad +Learned(TQA,Top1) &  $24.79_{\pm.21}$ & $\mathbf{24.71_{\pm.21}}$  \\
\midrule
BART-large (406M) & ${23.98_{\pm.20}}$ & ${24.28_{\pm.51}}$ \\
\quad +SSM & ${26.18_{\pm.24}}$ & $26.29_{\pm.43}$  \\
\quad +Learned(TQA,Top1) & $\mathbf{26.86_{\pm.08}}$ & $\mathbf{27.18_{\pm.34}}$ \\
\bottomrule
\end{tabular}
}
\caption{Performance Comparison on TriviaQA.}\label{tab:closed-book-tqa}
\vspace{-0.2cm}
\end{table}

\paragraph{Extending to More Datasets.} We further extend our experiments to policies learned from different anchor tasks and fine-tune across all three downstream task separately. We list the full results in Table \ref{tab:closed-book}. We observe that the learned masking policy ``transfers'' in some (but not all) cases: The Learned(NQ, Top5) policy helps to achieve $23.73\%_{(\pm.21)}$ EM on TriviaQA, which is on par with SSM; The Learned(TQA, Top5) transfers to Web Questions, also achieving results on par with SSM. This observation suggests that the learned policy may capture transferable meta-knowledge on ``what information is important for closed-book QA'', though such improvement is not consistent across all settings.

\paragraph{Discussion on NQ.} In Table \ref{tab:closed-book} we also observe that performances on NQ are close for all BART-base models; therefore it is hard to rank all compared methods. We argue that multiple factors leads to this phenomenon, including dataset characteristics and evaluation protocol.
Specifically, NQ may not be an ideal testbed for our study due to three reasons. 

Firstly, intermediate pre-training in general might not be as beneficial for this particular task. For instance,
\citet{roberts2020much} reports only 2\% EM gain on NQ using T5-11B. In our experiments, we use significantly smaller pre-trained models (BART-base/large), so the effect brought by intermediate pre-training will be even smaller. 
In our case we believe the effect is hidden in the variance brought by random seeds. 

Secondly, performance on NQ may not represent the real implicit knowledge capacity of a LM. For reference, we observe a 20\% dev set EM when fine-tuning a randomly initialized BART-base model on NQ. The general pre-training stage brings merely 4-5\% EM improvement, and therefore the improvement brought by intermediate pre-training can be marginal. 

And finally, evaluation based on exact match may substantially underestimate the model capability, as suggested in \cite{roberts2020much}.

\begin{table}[t]

    \centering
    \scalebox{0.6}{
    \begin{tabular}{p{10cm}}
    \toprule
        \makecell[l{p{10cm}}]{
        \textbf{Context:} \colorbox{yellow}{Jann Simon Wenner} (born \colorbox{yellow}{January 7, 1946}) is the co-founder and publisher of the popular culture biweekly magazine ``\colorbox{yellow}{Rolling Stone}'', and former owner of ``Men's Journal'' magazine. Born in \colorbox{yellow}{New York} City, Wenner graduated from \colorbox{yellow}{Chadwick School and later attended the University of California}, Berkeley.\\
        \textbf{Question:} What US popular culture magazine title, founded 1967 by Jann Wenner, was inspired by three so-named 1950s/60s musical sources?\\
        \textbf{Answer:} Rolling Stone}  \\
    \midrule
         \makecell[l{p{10cm}}]{
        \textbf{Context:} \colorbox{yellow}{Cape Fear is a 1991} American psychological thriller film directed by \colorbox{yellow}{Martin Scorsese and a remake of the 1962} film of the same name. It stars \colorbox{yellow}{Robert De Niro}, \colorbox{yellow}{Nick Nolte}, Jessica Lange, Joe Don Baker, Juliette Lewis, Robert Mitchum, and Gregory Peck in his final film role.\\
        \textbf{Question:} Gregory Peck, Robert Mitchum and Martin Balsam had cameo roles in the 1991 remake of what 1962 film in which they starred?\\
        \textbf{Answer:} Cape Fear}  \\   
    \bottomrule
    \end{tabular}
    }
    \vspace{-0.2cm}
    \caption{\textbf{Predictions of the learned policy on TriviaQA development set.} Top 10 spans predicted by the model are highlighted (overlapping predictions are merged for simpler visualization). The learned policy helps to predict salient spans in the context, which are helpful in answering the related question.}
    \label{tab:example}
    \vspace{-0.4cm}
\end{table}
\begin{table*}[t]
\centering
\scalebox{0.75}{
\begin{tabular}{lcccccc}
\toprule
\multirow{2}{*}{Model (\#param)} & \multicolumn{2}{c}{TriviaQA}   & \multicolumn{2}{c}{Natural Questions}  & \multicolumn{2}{c}{Web Questions}  \\
& dev & test & dev & test& dev & test \\
\midrule
BART-base (140M)& $22.34_{\pm.31}$ & $21.82_{\pm.15}$  & $24.66_{\pm.07}$ & $23.73_{\pm.25}$ & $24.19_{\pm.42}$ & $26.23_{\pm.05}$    \\
\quad +Random &  $23.02_{\pm.24}$ & $22.93_{\pm.14}$  & $25.47_{\pm.18}$ & $24.64_{\pm.44}$ & $24.75_{\pm.89}$ & $27.25_{\pm.68}$    \\
\quad +SSM &$24.53_{\pm.14}$ & $23.62_{\pm.29}$&  $25.03_{\pm.12}$ & $24.80_{\pm.06}$ & $28.44_{\pm.16}$ & $28.17_{\pm.40}$     \\
\quad +Learned (TQA, Top1) & $24.79_{\pm.21}$ & $\mathbf{24.71_{\pm.21}}$&  $25.31_{\pm.06}$  & $24.58_{\pm.19}$ & $26.87_{\pm.48}$& $27.84_{\pm.03}$   \\
\quad +Learned (TQA, Top5) & $\mathbf{24.84_{\pm.31}}$ & $24.43_{\pm.09}$&  $25.55_{\pm.15}$ & $24.66_{\pm.22}$ & $27.33_{\pm.42}$ & $28.35_{\pm.73}$  \\
\quad +Learned (NQ, Top1) & $23.93_{\pm.12}$ & $23.48_{\pm.10}$ & $25.40_{\pm.12}$ & $24.58_{\pm.10}$ & $26.59_{\pm.28}$ & $27.43_{\pm.38}$    \\
\quad +Learned (NQ, Top5) & $24.02_{\pm.09}$ & $23.73_{\pm.21}$&  $25.35_{\pm.38}$ & $24.86_{\pm.28}$ & $25.85_{\pm.58}$ & $28.15_{\pm.05}$  \\
\quad +Learned (All, Top1) & $24.68_{\pm.26}$ & $24.36_{\pm.14}$ & $25.37_{\pm.12}$ & $24.55_{\pm.47}$& $26.69_{\pm.80}$ & $28.33_{\pm1.01}$   \\
\quad +Learned (All, Top5)  & $24.76_{\pm.06}$ & $24.23_{\pm.25}$ & $25.20_{\pm.22}$ & $24.29_{\pm.17}$ & $27.33_{\pm.16}$ & $28.12_{\pm.36}$  \\
\midrule
BART-large (406M) & ${23.98_{\pm.20}}$ & ${24.28_{\pm.51}}$ & $25.89_{\pm.13}$ & $24.72_{\pm.16}$ & $27.89_{\pm1.25}$ & ${28.82_{\pm.33}}$  \\
\quad +SSM & ${26.18_{\pm.24}}$ & $26.29_{\pm.43}$& $26.34_{\pm.24}$ & $25.34_{\pm.23}$  & $29.55_{\pm.85}$ & $29.79_{\pm.47}$   \\
\quad +Learned (TQA, Top1) & $\mathbf{26.86_{\pm.08}}$ & $\mathbf{27.18_{\pm.34}}$ &  $25.41_{\pm.26}$  & $24.28_{\pm.28}$   & $29.64_{\pm1.21}$ & $29.71_{\pm.74}$  \\
\bottomrule
\end{tabular}
}
\caption{\textbf{Exact match (EM) on Closed-book QA datasets.} ``+Learned (TQA, Top1)'' means we learn the masking policy from (context, answer) examples in TriviaQA, and mask the top 1 span during pre-training. We report 3-run average and standard deviation with different random seeds.}\label{tab:closed-book}
\end{table*}

\subsection{Case Study of the Learned Policy} 
We use the policy trained on TriviaQA and provide its prediction on TriviaQA dev set context paragraphs in Table \ref{tab:example}. We observe that the policy is masking important spans within the context paragraph as expected. One limitation is that the start position and end position are determined independently, so that ``Chadwick School and later attended the University of California'' becomes an candidate answer span because the start position logit for ``Chadwick'' is large, and the end position logit for ``California'' is large. Ideally ``Chadwick School'' and ``University of California'' should be masked separately. Similar problem holds for machine reading comprehension models, and this is a limitation of our current approach. 

\section{Related Work}


\paragraph{Implicit Knowledge in Pre-trained Language Models.} 
\citet{petroni-etal-2019-language} discovered that pre-trained language models can implicitly store relational knowledge in their parameters, and such knowledge can be accessed with cloze-style queries. \citet{roberts2020much} introduced the task of closed-book QA, which breaks the convention of two-stage retriever-reader strategy for open-domain QA, and requires the model to directly generate answers with its implicit knowledge. Closed-book QA performance is boosted significantly when salient span masking \cite{guu2020realm} is used. \citet{guu2020realm} maintained that SSM helps the model to ``focus on problems that require world knowledge''.

\paragraph{Self-supervised Pre-training.} 
Pre-trained language models has shown its capability on a wide variety of NLP tasks. Current self-supervised objectives are mostly \textit{heuristic}, including masked language modeling \cite{devlin-etal-2019-bert}, span boundary representation learning \cite{joshi-etal-2020-spanbert}, corrupted sentence reconstruction \cite{lewis-etal-2020-bart}, gap sentence generation \cite{Zhang2019PEGASUSPW}, etc. \citet{raffel2019exploring} systematically studied the self-supervised objectives used in previous literature. Related to our goal of exploring pre-training objectives, ELECTRA \cite{clark2020electra} propose a replaced token prediction task which improves pre-training efficiency. \citet{chen2020variance} propose to reduce the variance of gradients in SGD and expedite model pre-training. \citet{Levine2020pmimasking} propose to mask n-grams according to Pointwise Mutual Information (PMI). These works typically consider the efficiency of an objective when pre-training \textit{from scratch} and without preconceived focus on a given problem; while we focus on encoding implicit knowledge during intermediate pre-training with a given set of tasks in mind.

\paragraph{Domain/Task-specific Pre-training.}
\citet{gururangan-etal-2020-dont} experiment on 4 different domains (bio-medical, computer science, news, reviews) and 8 different datasets, where they discover that pre-training with in-domain corpus leads to better downstream performance. \citet{kang-etal-2020-neural} propose to learn a mask generator during task-specific pre-training via reinforcement learning. Closely related to us, \citet{gu-etal-2020-train} propose task-guided pre-training by first learning to predict importance score for individual tokens and then launch task-specific pre-training by masking important tokens. 
\section{Conclusion}
In this paper we propose a simple and intuitive method to encode task-relevant knowledge into language models during intermediate pre-training. Our method resembles how students would prepare for a closed-book exam: reading the books in advance (pre-training on Wikipedia), figuring out what contents would likely appear in the test, and focusing on that content (by masking task-relevant spans). We showed the learned policy is particularly effective on TriviaQA; meanwhile the trend is less clear on other datasets, for which we discussed the reasons behind. We also observe some cases where a policy learned on one QA dataset can positively transfer to help improve performance of another, suggesting that the meta-knowledge captured by the learned policy can be transferable. 

There are several potential directions for future work. Firstly, what the masking policy captures is closely related to the concept of ``learning to learn'', and thus future work may exploit meta-learning methods to learn the masking policy. Secondly, given recent advances in unsupervised QA \cite{lewis-etal-2019-unsupervised, shwartz-etal-2020-unsupervised}, it would be interesting to enable self-training by forming a closed-loop system that learns to memorize knowledge, ask questions, and provide answers.

\paragraph{Acknowledgments} 
We would like to thank Sebastian Riedel, Edward Grefenstette, Patrick Lewis, and Xian Li for the insightful discussions. Qinyuan Ye's research is supported in part by the Office of the Director of National Intelligence (ODNI), Intelligence Advanced Research Projects Activity (IARPA), via Contract No. 2019-19051600007.



\bibliography{anthology,acl2020}
\bibliographystyle{acl_natbib}

\appendix

\section{Hyperparameters}
For closed-book QA fine-tuning, we first select the learning rate from $\{$5e-6, 1e-5, 2e-5, 3e-5$\}$ and then fix learning rate to select batch size from $\{$32, 64, 128, 256$\}$. 

\label{app:hps}
\begin{table}[h]
\centering
\scalebox{0.7}{
\begin{tabular}{ll}
\toprule
Parameter Name            & Value                          \\
\midrule
Max Epoch                 & 100                            \\
Validation Interval       & 2 or 5                         \\
Warmup Updates            & 500                            \\
Learning Rate             & $\{$5e-6, 1e-5, 2e-5, 3e-5$\}$ \\
Batch Size                & $\{$32, 64, 128, 256$\}$       \\
Label Smoothing           & 0.1                            \\
Dropout                   & 0.1                            \\
Weight Decay              & 0.01                           \\
Clip Norm                 & 0.1                            \\
Generation Beam Size      & 4                              \\
Generation Min/Max Length & 1/20                           \\
Generation Length Penalty & 1.0                           \\
\bottomrule
\end{tabular}
}
\caption{Hyperparameters for fine-tuning on closed-book QA tasks.}
\end{table}

\end{document}